\documentclass[10pt,twocolumn,letterpaper]{article}

\usepackage{iccv}
\usepackage{times}
\usepackage{epsfig}
\usepackage{graphicx}
\usepackage{amsmath}
\usepackage{amssymb}
\usepackage{xcolor}
\usepackage{multirow}
\usepackage{multicol}
\usepackage{booktabs}
\usepackage{subcaption}
\usepackage{xspace}
\usepackage{pifont}
\usepackage{graphicx,verbatimbox}
\usepackage{multirow, booktabs}

\usepackage[pagebackref=true,breaklinks=true,colorlinks,bookmarks=false]{hyperref}
\usepackage[british,UKenglish,USenglish,english,american]{babel}

\newcommand{\cmark}{\ding{51}\xspace}%

\iccvfinalcopy %

\begin{document}

\title{CvT: Introducing Convolutions to Vision Transformers}

\author{
Haiping Wu$^{1,2}$$\thanks{This work is done when Haiping Wu was an intern at Microsoft.}$
\and Bin Xiao$^{2}$$\thanks{Corresponding author}$
\and Noel Codella$^{2}$ 
\and Mengchen Liu$^{2}$ 
\and Xiyang Dai$^{2}$ 
\and Lu Yuan$^{2}$
~~ Lei Zhang$^{2}$
\\
$^{1}$McGill University \quad\quad\quad\quad\quad\quad$^{2}$Microsoft Cloud + AI~~~\\
{
\tt\small haiping.wu2@mail.mcgill.ca,
\tt\small \{bixi, ncodella, mengcliu, xidai, luyuan, leizhang\}@microsoft.com
}
}

\maketitle


\begin{abstract}
We present in this paper a new architecture, named Convolutional vision Transformer (CvT), that improves Vision Transformer (ViT) in performance and efficiency by introducing convolutions into ViT to yield the best of both designs. This is accomplished through two primary modifications: a hierarchy of Transformers containing a new convolutional token embedding, and a convolutional Transformer block leveraging a convolutional projection. These changes introduce desirable properties of convolutional neural networks (CNNs) to the ViT architecture (\ie shift, scale, and distortion invariance) while maintaining the merits of Transformers (\ie dynamic attention, global context, and better generalization). We validate CvT by conducting extensive experiments, showing that this approach achieves state-of-the-art performance over other Vision Transformers and ResNets on ImageNet-1k, with fewer parameters and lower FLOPs. In addition, performance gains are maintained when pretrained on larger datasets (\eg ImageNet-22k) and fine-tuned to downstream tasks. Pre-trained on ImageNet-22k, our CvT-W24 obtains a top-1 accuracy of 87.7\% on the ImageNet-1k val set. Finally, our results show that the positional encoding, a crucial component in existing Vision Transformers, can be safely removed in our model, simplifying the design for higher resolution vision tasks. Code will be released at \url{https://github.com/leoxiaobin/CvT}.
\end{abstract}

\section{Introduction}

\begin{figure}[ht]
    \centering
    \includegraphics[width=1.0\linewidth]{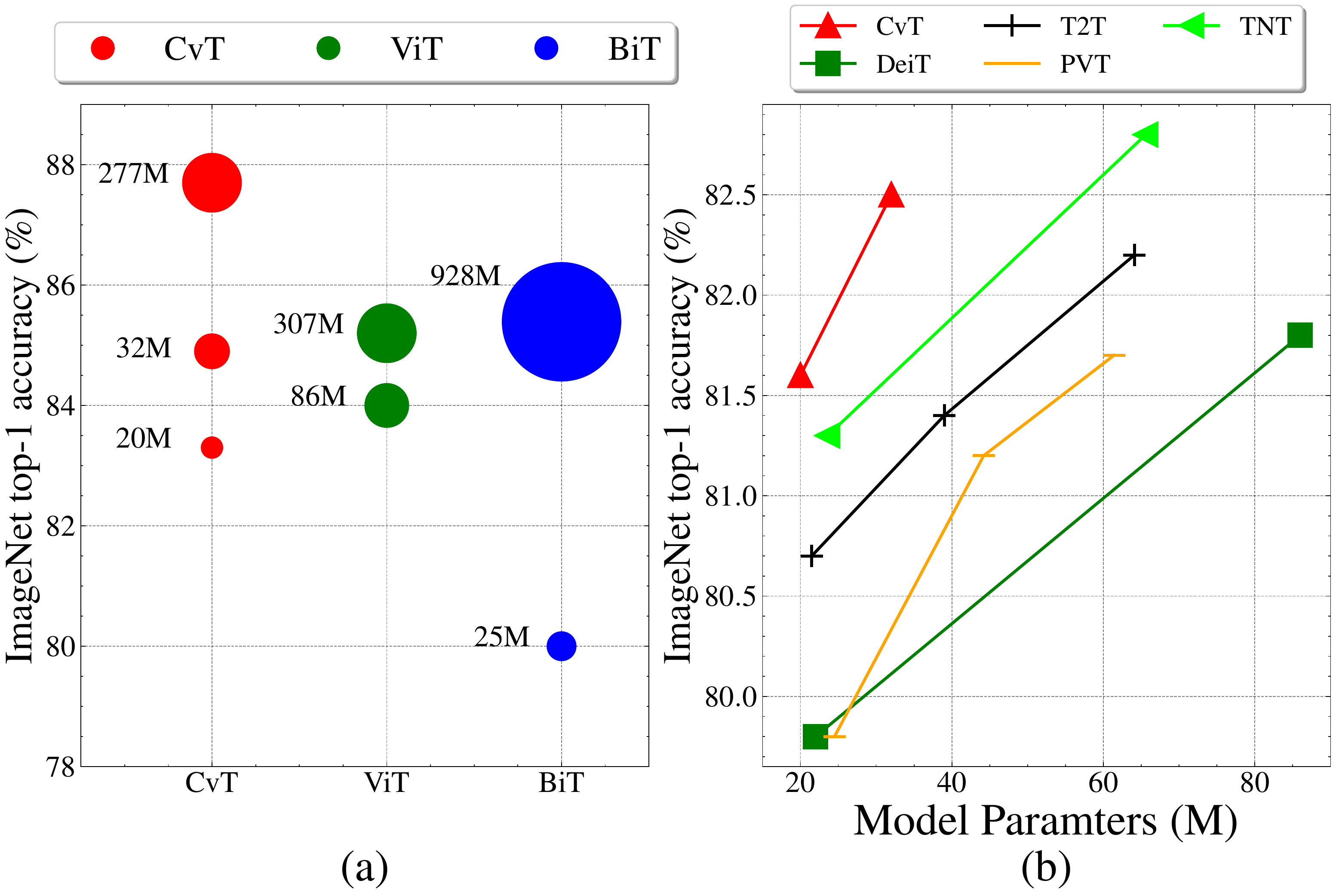}
    \caption{Top-1 Accuracy on ImageNet validation compared to other methods with respect to model parameters. 
    (a) Comparison to CNN-based model BiT~\cite{kolesnikov2019big} and Transformer-based model ViT~\cite{dosovitskiy2020image}, when pretrained on ImageNet-22k. Larger marker size indicates larger architectures. (b) Comparison to concurrent works: DeiT~\cite{touvron2020training}, T2T~\cite{yuan2021tokens}, PVT~\cite{wang2021pyramid}, TNT~\cite{han2021transformer} when pretrained on ImageNet-1k. }
    \label{fig:model_compare}
\end{figure}

\begin{table*}[ht]
\centering
\setlength{\tabcolsep}{2.0pt}
    \small
    \begin{tabular}{c|c|c|c|c}
\toprule[1.2pt]
~ Method ~ & Needs Position Encoding (PE) & ~ Token Embedding ~ & ~ Projection for Attention ~ & Hierarchical Transformers \\
\hline\hline
ViT~\cite{dosovitskiy2020image}, DeiT~\cite{touvron2020training} & yes  & non-overlapping & linear & no \\
\hline
CPVT~\cite{chu2021really} & no (w/ PE Generator)  & non-overlapping & linear & no \\
\hline
TNT~\cite{han2021transformer} & yes  & non-overlapping (patch+pixel) & linear & no \\
\hline
T2T~\cite{yuan2021tokens} & yes  & overlapping (concatenate) & linear & partial (tokenization) \\
\hline
PVT~\cite{wang2021pyramid} & yes  & non-overlapping & spatial reduction & yes \\
\hline
CvT (ours) & no  & overlapping (convolution)& convolution & yes \\
\hline
\end{tabular} \vspace{-0.5em} \vspace{-0.7em} 
\caption{Representative works of vision Transformers.}
\label{tb:relatedwork}
\end{table*}

Transformers ~\cite{vaswani2017attention,devlin2018bert} have recently dominated a wide range of tasks in natural language processing (NLP)~\cite{wang2018glue}. The Vision Transformer (ViT)~\cite{dosovitskiy2020image} is the first computer vision model to rely exclusively on the Transformer architecture to obtain competitive image classification performance at large scale. The ViT design adapts Transformer architectures~\cite{devlin2018bert} from language understanding with minimal modifications. First, images are split into discrete non-overlapping patches (\eg $16 \times 16$). Then, these patches are treated as tokens (analogous to tokens in NLP), summed with a special positional encoding to represent coarse spatial information, and input into repeated standard Transformer layers to model global relations for classification. 

Despite the success of vision Transformers at large scale, the performance is still below similarly sized convolutional neural network (CNN) counterparts (\eg, ResNets~\cite{he2016deep}) when trained on smaller amounts of data. One possible reason may be that ViT lacks certain desirable properties inherently built into the CNN architecture that make CNNs uniquely suited to solve vision tasks. For example, images have a strong 2D local structure: spatially neighboring pixels are usually highly correlated. The CNN architecture forces the capture of this local structure by using {\em local receptive fields, shared weights, and spatial subsampling}~\cite{Lecun99objectrecognition}, and thus also achieves some degree of shift, scale, and distortion invariance. In addition, the hierarchical structure of convolutional kernels learns visual patterns that take into account local spatial context at varying levels of complexity, from simple low-level edges and textures to higher order semantic patterns.

In this paper, we hypothesize that convolutions can be strategically introduced to the ViT structure to improve performance and robustness, while concurrently maintaining a high degree of computational and memory efficiency. To verify our hypothesises, we present a new architecture, called the Convolutional vision Transformer (CvT), which incorporates convolutions into the Transformer that is inherently efficient, both in terms of floating point operations (FLOPs) and parameters.

The CvT design introduces convolutions to two core sections of the ViT architecture. First, we partition the Transformers into \textit{multiple stages} that form a hierarchical structure of Transformers. The beginning of each stage consists of a \textit{convolutional token embedding} that performs an overlapping convolution operation with stride on a 2D-reshaped token map (\ie, reshaping flattened token sequences back to the spatial grid), followed by layer normalization. This allows the model to not only capture local information, but also progressively decrease the sequence length while simultaneously increasing the dimension of token features across stages, achieving spatial downsampling while concurrently increasing the number of feature maps, as is performed in CNNs~\cite{Lecun99objectrecognition}. 
Second, the linear projection prior to every self-attention block in the Transformer module is replaced with our proposed {\em convolutional projection}, which employs a $s \times s $ depth-wise separable convolution~\cite{chollet2017xception} operation on an 2D-reshaped token map. This allows the model to further capture local spatial context and reduce semantic ambiguity in the attention mechanism. It also permits management of computational complexity, as the stride of convolution can be used to subsample the key and value matrices to improve efficiency by 4$\times$ or more, with minimal degradation of performance.

In summary, our proposed Convolutional vision Transformer (CvT) employs all the benefits of CNNs: {\em local receptive fields, shared weights, and spatial subsampling}, while keeping all the advantages of Transformers: {\em dynamic attention, global context fusion, and better generalization}. Our results demonstrate that this approach attains state-of-art performance when CvT is pre-trained with ImageNet-1k, while being lightweight and efficient: CvT improves the performance compared to CNN-based models (\eg ResNet) and prior Transformer-based models (\eg ViT, DeiT) while utilizing fewer FLOPS and parameters. In addition, CvT achieves state-of-the-art performance when evaluated at larger scale pretraining (\eg on the public ImageNet-22k dataset). Finally, we demonstrate that in this new design, we can drop the positional embedding for tokens without any degradation to model performance. This not only simplifies the architecture design, but also makes it readily capable of accommodating variable resolutions of input images that is critical to many vision tasks.

\begin{figure*}[!ht]
    \centering
    \includegraphics[width=\textwidth]{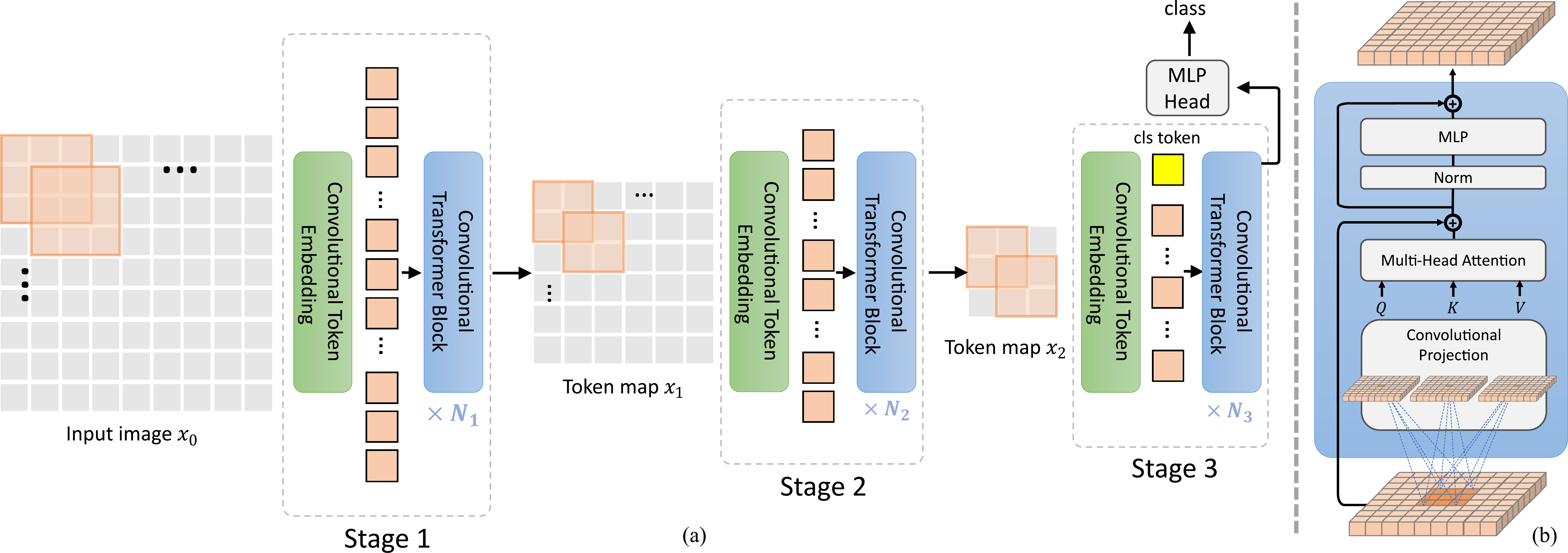}
    \caption{The pipeline of the proposed CvT architecture. (a) Overall architecture, showing the hierarchical multi-stage structure facilitated by the Convolutional Token Embedding layer. (b) Details of the Convolutional Transformer Block, which contains the convolution projection as the first layer.}
    \label{fig:pipeline}
\end{figure*}

\section{Related Work}

Transformers that exclusively rely on the self-attention mechanism to capture global dependencies have dominated in natural language modelling~\cite{vaswani2017attention, devlin2018bert, radford2018improving}. Recently, the Transformer based architecture has been viewed as a viable alternative to the convolutional neural networks (CNNs) in visual recognition tasks, such as classification~\cite{dosovitskiy2020image, touvron2020training}, object detection~\cite{carion2020end,zhu2020deformable,zheng2020end,dai2020up,sun2020rethinking}, segmentation~\cite{wang2020max,wang2020end}, 
image enhancement~\cite{chen2020pre,yang2020learning}, image generation~\cite{parmar2018image}, video processing~\cite{zeng2020learning,Zhou_2018_CVPR} and 3D point cloud processing~\cite{engel2020point}. \vspace{-0.5em}

\paragraph{Vision Transformers.}

The Vision Transformer (ViT) is the first to prove that a pure Transformer architecture can attain state-of-the-art
performance (\eg ResNets~\cite{he2016deep}, EfficientNet~\cite{tan2019efficientnet}) on image classification when the data is large enough (\ie on ImageNet-22k, JFT-300M). Specifically, ViT decomposes each image into a sequence of tokens (\ie non-overlapping patches) with fixed length, and then applies multiple standard Transformer layers, consisting of Multi-Head Self-Attention module (MHSA) and Position-wise Feed-forward module (FFN), to model these tokens. DeiT~\cite{touvron2020training} further explores the data-efficient training and distillation for ViT. In this work, we study how to combine CNNs and Transformers to model both local and global dependencies for image classification in an efficient way. 

In order to better model local context in vision Transformers, some concurrent works have introduced design changes. For example, the Conditional Position encodings Visual Transformer (CPVT)~\cite{chu2021really}  replaces the predefined positional embedding used in ViT with conditional position encodings (CPE), enabling Transformers to process input images of arbitrary size without interpolation. Transformer-iN-Transformer (TNT)~\cite{han2021transformer} utilizes both an outer Transformer block that processes the patch embeddings, and an inner Transformer block that models the relation among pixel embeddings, to model both patch-level and pixel-level representation. Tokens-to-Token (T2T)~\cite{yuan2021tokens} mainly improves tokenization in ViT by concatenating multiple tokens within a sliding window into one token. However, this operation fundamentally differs from convolutions especially in normalization details, and the concatenation of multiple tokens greatly increases complexity in computation and memory. PVT~\cite{wang2021pyramid} incorporates a multi-stage design (without convolutions) for Transformer similar to multi-scales in CNNs, favoring dense prediction tasks. 

In contrast to these concurrent works, this work aims to achieve the best of both worlds by introducing convolutions, with image domain specific inductive biases, into the Transformer architecture. Table~\ref{tb:relatedwork} shows the key differences in terms of {\em necessity of positional encodings, type of token embedding, type of projection}, and {\em Transformer structure in the backbone}, between the above representative concurrent works and ours. \vspace{-0.5em}

\paragraph{Introducing Self-attentions to CNNs.}

Self-attention mechanisms have been widely applied to CNNs in vision tasks. Among these works, the non-local networks~\cite{wang2018non} are designed for capturing long range dependencies via global attention. The local relation networks~\cite{hu2019local} adapts its weight aggregation based on the compositional relations (similarity) between pixels/features within a local window, in contrast to convolution layers which employ fixed aggregation weights over spatially neighboring input feature. Such an adaptive weight aggregation introduces geometric priors into the network which are important for the recognition tasks. Recently, BoTNet~\cite{srinivas2021bottleneck} proposes a simple yet powerful backbone architecture that just replaces the spatial convolutions with global self-attention in the final three bottleneck blocks of a ResNet and achieves a strong performance in image recognition. Instead, our work performs an opposite research direction: introducing convolutions to Transformers. \vspace{-0.5em}

\paragraph{Introducing Convolutions to Transformers.}

In NLP and speech recognition, convolutions have been used to modify the Transformer block, either by replacing multi-head attentions with convolution layers~\cite{wu2019pay}, or adding additional convolution layers in parallel ~\cite{wu2020lite} or sequentially~\cite{gulati2020conformer}, to capture local relationships. Other prior work ~\cite{wang2021evolving} proposes to propagate attention maps to succeeding layers via a residual connection, which is first transformed by convolutions. Different from these works, we propose to introduce convolutions to two primary parts of the vision Transformer: first, to replace the existing Position-wise Linear Projection for the attention operation with our Convolutional Projection, and second, to use our hierarchical multi-stage structure to enable varied resolution of 2D re-shaped token maps, similar to CNNs. Our unique design affords significant performance and efficiency benefits over prior works.

\section{Convolutional vision Transformer}

The overall pipeline of the Convolutional vision Transformer (CvT) is shown in Figure~\ref{fig:pipeline}. We introduce two convolution-based operations into the Vision Transformer architecture, namely the ~\textit{Convolutional Token Embedding} and ~\textit{Convolutional Projection}.
As shown in Figure~\ref{fig:pipeline} (a), a multi-stage hierarchy design borrowed from CNNs~\cite{Lecun99objectrecognition, he2016deep} is employed, where three stages in total are used in this work. Each stage has two parts. First, the input image (or 2D reshaped token maps) are subjected to the \emph{Convolutional Token Embedding} layer, which is implemented as a convolution with overlapping patches with tokens reshaped to the 2D spatial grid as the input (the degree of overlap can be controlled via the stride length). An additional layer normalization is applied to the tokens. 
This allows each stage to progressively reduce the number of tokens (\ie feature resolution) while simultaneously increasing the width of the tokens (\ie feature dimension), thus achieving spatial downsampling and increased richness of representation, similar to the design of CNNs. 
Different from other prior Transformer-based architectures~\cite{dosovitskiy2020image, touvron2020training, yuan2021tokens, wang2021pyramid}, we do not sum the ad-hod position embedding to the tokens.
Next, a stack of the proposed \emph{Convolutional Transformer Blocks} comprise the remainder of each stage. Figure~\ref{fig:pipeline} (b) shows the architecture of the Convolutional Transformer Block, where a depth-wise separable convolution operation~\cite{chollet2017xception}, referred as \emph{Convolutional Projection}, is applied for query, key, and value embeddings respectively, instead of the standard position-wise linear projection in ViT~\cite{dosovitskiy2020image}. 
Additionally, the classification token is added only in the last stage.
Finally, an MLP (\ie fully connected) Head is utilized upon the classification token of the final stage output to predict the class.

We first elaborate on the proposed \emph{Convolutional Token Embedding} layer. Next we show how to perform \emph{Convolutional Projection} for the Multi-Head Self-Attention module, and its efficient design for managing computational cost.

\subsection{Convolutional Token Embedding}

This convolution operation in CvT aims to model local spatial contexts, from low-level edges to higher order semantic primitives, over a multi-stage hierarchy approach, similar to CNNs.

Formally, given a 2D image or a 2D-reshaped output token map from a previous stage $x_{i-1}\in\mathbb{R}^{H_{i-1} \times W_{i-1} \times C_{i-1}}$ as the input to stage $i$, 
we learn a function $f(\cdot)$ that maps $x_{i-1}$ into new tokens $f(x_{i-1})$ with a channel size $C_i$, where $f(\cdot)$ is 2D convolution operation of kernel size $s\times s$, stride $s-o$ and $p$ padding (to deal with boundary conditions). The new token map $f(x_{i-1})\in \mathbb{R}^{H_{i} \times W_{i} \times C_i}$ has height and width 
\begin{equation}
    H_{i} = \left\lfloor\frac{H_{i-1}+ 2p-s}{s-o}+1\right\rfloor,
    W_{i} = \left\lfloor\frac{W_{i-1}+ 2p-s}{s-o}+1\right\rfloor.
\end{equation}
$f(x_{i-1})$ is then flattened into size $H_iW_i \times C_i$ and normalized by layer normalization~\cite{ba2016layer} for input into the subsequent Transformer blocks of stage $i$. 

The \emph{Convolutional Token Embedding} layer allows us to adjust the token feature dimension and the number of tokens at each stage by varying parameters of the convolution operation. In this manner, in each stage we progressively decrease the token sequence length, while increasing the token feature dimension. This gives the tokens the ability to represent increasingly complex visual patterns over increasingly larger spatial footprints, similar to feature layers of CNNs.

\subsection{Convolutional Projection for Attention}
\begin{figure*}[ht]
    \centering
    \includegraphics[width=\textwidth]{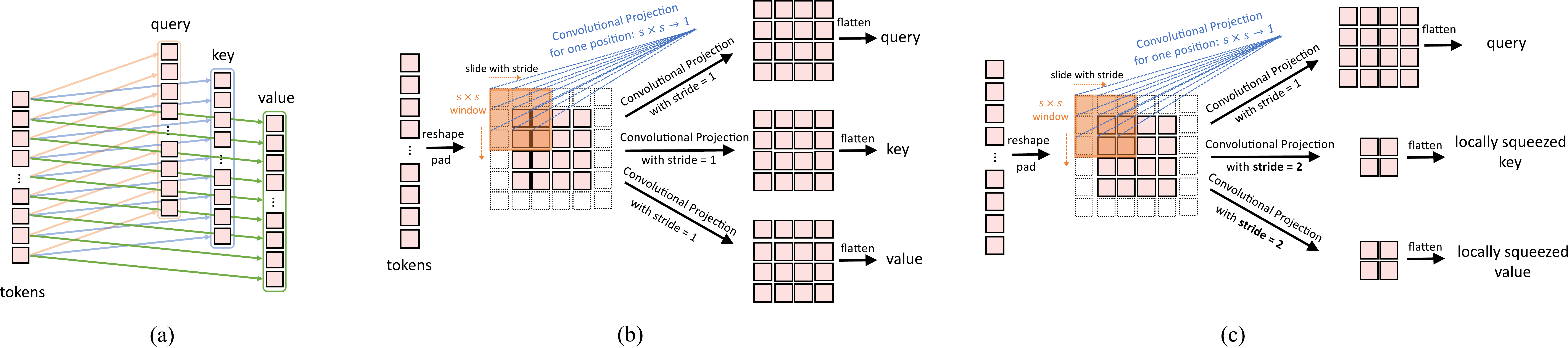}\vspace{-0.3em}
    \caption{(a) Linear projection in ViT~\cite{dosovitskiy2020image}. (b) Convolutional projection. (c) Squeezed convolutional projection. Unless otherwise stated, we use (c) Squeezed convolutional projection by default.
    }
    \label{fig:qkv_proj}
\end{figure*}

 The goal of the proposed \emph{Convolutional Projection} layer is to achieve additional modeling of local spatial context, and to provide efficiency benefits by permitting the undersampling of $K$ and $V$ matrices. 
 
 Fundamentally, the proposed Transformer block with \emph{Convolutional Projection} is a generalization of the original Transformer block. While previous works~\cite{gulati2020conformer, wu2020lite} try to add additional convolution modules to the Transformer Block for speech recognition and natural language processing, they result in a more complicated design and additional computational cost. Instead, we propose to replace the original position-wise linear projection for Multi-Head Self-Attention (MHSA) with depth-wise separable convolutions, forming the ~\emph{Convolutional Projection} layer.  

\subsubsection{Implementation Details}

Figure~\ref{fig:qkv_proj} (a) shows the 
original position-wise linear projection used in ViT~\cite{dosovitskiy2020image} and Figure~\ref{fig:qkv_proj} (b) shows our proposed $s\times s$ \emph{Convolutional Projection}. As shown in Figure~\ref{fig:qkv_proj} (b), tokens are first reshaped into 
a 2D token map. Next, a Convolutional Projection is implemented using a depth-wise separable convolution layer with kernel size $s$. Finally, the projected tokens are flattened into 1D for subsequent process. This can be formulated as: 
 \begin{equation}
     x_i^{q/k/v} = \texttt{Flatten}\left(\texttt{Conv2d}\left(\texttt{Reshape2D}(x_i), \; s\right)\right),
 \end{equation}
 where $x_i^{q/k/v}$ is the token input for $Q/K/V$ matrices at layer $i$, $x_i$ is the unperturbed token prior to the Convolutional Projection, $\texttt{Conv2d}$ is a depth-wise separable convolution~\cite{chollet2017xception} implemented by: $\texttt{Depth-wise~Conv2d} \rightarrow \texttt{BatchNorm2d} \rightarrow$ $\texttt{Point-wise~Conv2d}$, and $s$ refers to the convolution kernel size.

The resulting new Transformer Block with the Convolutional Projection layer is a generalization of the original Transformer Block design. The original position-wise linear projection layer could be trivially implemented using a convolution layer with kernel size of $1 \times 1$.

\subsubsection{Efficiency Considerations}
There are two primary efficiency benefits from the design of our Convolutional Projection layer. 

First, we utilize efficient convolutions. Directly using standard $s \times s$ convolutions for the Convolutional Projection would require $s^2C^2$ parameters and $\mathcal{O}(s^2C^2T)$ FLOPs,
where $C$ is the token channel dimension, and $T$ is the number of tokens for processing. Instead, we split the standard $s \times s$ convolution into a depth-wise separable convolution ~\cite{howard2017mobilenets}. In this way, each of the proposed Convolutional Projection would only introduce an extra of $s^2C$ parameters and $\mathcal{O}(s^2CT)$ FLOPs compared to the original position-wise linear projection, which are negligible with respect to the total parameters and FLOPs of the models.

Second, we leverage the proposed Convolutional Projection to reduce the computation cost for the MHSA operation. The $s \times s$ Convolutional Projection permits reducing the number of tokens by using a stride larger than 1. 
Figure~\ref{fig:qkv_proj} (c) shows the Convolutional Projection, where the key and value projection are subsampled by using a convolution with stride larger than 1. 
We use a stride of 2 for key and value projection, leaving the stride of 1 for query unchanged. In this way, the number of tokens for key and value is reduced 4 times, and the computational cost is reduced by 4 times for the later MHSA operation. This comes with a minimal performance penalty, as neighboring pixels/patches in images tend to have redundancy in appearance/semantics. In addition, the local context modeling of the proposed Convolutional Projection compensates for the loss of information incurred by resolution reduction.

\begin{table*}[t]
\small
    \centering
    \scalebox{0.92}
    {
    \begin{tabular}{c|c|l|c|c|c}
\hline
\multicolumn{1}{l|}{}     
& Output Size 
& Layer Name                                                          
& CvT-13 
& CvT-21 
& CvT-W24 \\ 
\hline

\multirow{4}{*}{Stage1}    
& $56\times56$
& Conv. Embed.                                                     
& \multicolumn{2}{c|}{$7\times7$, $64$, stride $4$}                   
& $7\times7$, $192$, stride $4$               \\ 
\cline{2-6} 

& $56\times56$
& \begin{tabular}[c]{@{}l@{}}Conv. Proj.\\MHSA\\ MLP\end{tabular} 
&      
$\begin{bmatrix}
\begin{array}{c}
3\times3, 64\\
H_1=1, D_1=64\\
R_1=4 
\end{array}
\end{bmatrix} \times 1$
&             
$\begin{bmatrix}
\begin{array}{c}
     3\times3, 64\\
     H_1=1, D_1=64\\
     R_1=4
\end{array}
\end{bmatrix} \times 1$
&      
$\begin{bmatrix}
\begin{array}{c}
     3\times3, 192\\
     H_1=3, D_1=192\\
     R_1=4  
\end{array}
\end{bmatrix} \times 2$
\\ \hline

\multirow{4}{*}{Stage2}    
& $28\times28$
&  Conv. Embed.
& \multicolumn{2}{c|}{$3\times3$, $192$, stride $2$}                   
& $3\times3$, $768$, stride $2$
\\ \cline{2-6} 
                           
& $28\times28$
& \begin{tabular}[c]{@{}l@{}}Conv. Proj.\\MHSA\\ MLP\end{tabular} 
&             
$\begin{bmatrix}
\begin{array}{c}
     3\times3, 192\\
     H_2=3, D_2=192\\
     R_2=4  
\end{array}
\end{bmatrix} \times 2$
&             
$\begin{bmatrix}
\begin{array}{c}
     3\times3, 192\\
     H_2=3, D_2=192\\
     R_2=4 
\end{array}
\end{bmatrix} \times 4$
&             
$\begin{bmatrix}
\begin{array}{c}
     3\times3, 768\\
     H_2=12, D_2=768\\
     R_2=4  
\end{array}
\end{bmatrix} \times 2$
\\ \hline

\multirow{4}{*}{Stage3}    
& $14\times14$            
& Conv. Embed.
& \multicolumn{2}{c|}{$3\times3$, $384$, stride $2$}                   
& $3\times3$, $1024$, stride $2$
\\ \cline{2-6} 
                           
&$14\times14$ 
& \begin{tabular}[c]{@{}l@{}}Conv. Proj.\\MHSA\\ MLP\end{tabular} 
&
$\begin{bmatrix}
\begin{array}{c}
     3\times3, 384\\
     H_3=6, D_3=384\\
     R_3=4 
\end{array}
\end{bmatrix} \times 10$
&             
$\begin{bmatrix}
\begin{array}{c}
     3\times3, 384\\
     H_3=6, D_3=384\\
     R_3=4  
\end{array}
\end{bmatrix} \times 16$
&
$\begin{bmatrix}
\begin{array}{c}
     3\times3, 1024\\
     H_3=16, D_3=1024\\
     R_3=4 
\end{array}
\end{bmatrix} \times 20$
\\ \hline

\multicolumn{1}{c|}{Head} 
& $1\times1$
& Linear
& \multicolumn{3}{c}{1000}
\\ \hline

\multicolumn{3}{c|}{Params}  
& $19.98$ M 
& $31.54$ M
& $276.7$ M
\\ \hline
\multicolumn{3}{c|}{FLOPs}  
& $4.53$ G
& $7.13$ G
& $60.86$ G
\\ \hline
\end{tabular}}\vspace{-0.3em}
    \caption{Architectures for ImageNet classification. Input image size is $224\times224$ by default. Conv. Embed.: Convolutional Token Embedding. Conv. Proj.: Convolutional Projection. $H_i$ and $D_i$ is the number of  heads and embedding feature dimension in the $i$th MHSA module. $R_i$ is the feature dimension expansion ratio in the $i$th MLP layer.}
    \label{tab:arch}
\end{table*}

\subsection{Methodological Discussions}

\paragraph{Removing Positional Embeddings:}
The introduction of \emph{Convolutional Projections} for every Transformer block, combined with the \emph{Convolutional Token Embedding}, gives us the ability to model local spatial relationships through the network. This built-in property allows dropping the position embedding from the network without hurting performance, as evidenced by our experiments (Section{\color{red}{~\ref{abl:pe}}}), simplifying design for vision tasks with variable input resolution. \vspace{-0.5em}

\paragraph{Relations to Concurrent Work:}
Recently, two more related concurrent works also propose to improve ViT by incorporating elements of CNNs to Transformers. Tokens-to-Token ViT~\cite{yuan2021tokens} implements a progressive tokenization, and then uses a Transformer-based backbone in which the length of tokens is fixed. By contrast, our CvT implements a progressive tokenization by a multi-stage process -- containing both convolutional token embeddings and convolutional Transformer blocks in each stage. As the length of tokens are decreased in each stage, the width of the tokens (dimension of feature) can be increased, allowing increased richness of representations at each feature spatial resolution. Additionally, whereas T2T concatenates neighboring tokens into one new token, leading to increasing the complexity of memory and computation, our usage of convolutional token embedding directly performs contextual learning without concatenation, while providing the flexibility of controlling stride and feature dimension. To manage the complexity, T2T has to consider a deep-narrow architecture design with smaller hidden dimensions and MLP size than ViT in the subsequent backbone. Instead, we changed previous Transformer modules by replacing the position-wise linear projection with our convolutional projection

Pyramid Vision Transformer (PVT)~\cite{wang2021pyramid} overcomes the difficulties of porting ViT to various dense prediction tasks. In ViT, the output feature map has only a single scale with low resolution. In addition, computations and memory cost are relatively high, even for common input image sizes. To address this problem, both PVT and our CvT incorporate pyramid structures from CNNs to the Transformers structure. Compared with PVT, which only spatially subsamples the feature map or key/value matrices in projection, our CvT instead employs convolutions with stride to achieve this goal. Our experiments (shown in Section{\color{red}{~\ref{abl:conv_proj}}}) demonstrate that the fusion of local neighboring information plays an important role on the performance.

\section{Experiments}
In this section, we evaluate the CvT model on large-scale image classification datasets and transfer to various downstream datasets. In addition, we perform through ablation studies to validate the design of the proposed architecture.

\subsection{Setup}
For evaluation, we use the ImageNet dataset, with 1.3M images and 1k classes, as well as its superset ImageNet-22k with 22k classes and 14M images~\cite{deng2009imagenet}. We further transfer the models pretrained on ImageNet-22k to downstream tasks, including CIFAR-10/100~\cite{krizhevsky2009learning}, Oxford-IIIT-Pet~\cite{parkhi12a}, Oxford-IIIT-Flower~\cite{Nilsback08}, following~\cite{kolesnikov2019big, dosovitskiy2020image}. \vspace{-0.5em}

\paragraph{Model Variants}
We instantiate models with different parameters and FLOPs by varying the number of Transformer blocks of each stage and the hidden feature dimension used, as shown in Table~\ref{tab:arch}. Three stages are adapted. 
We define CvT-13 and CvT-21 as basic models, with 19.98M and 31.54M paramters. CvT-$X$ stands for Convolutional vision Transformer with $X$ Transformer Blocks in total. Additionally, we experiment with a wider model with a larger token dimension for each stage, namely CvT-W24 (W stands for Wide), resulting 298.3M parameters, to validate the scaling ability of the proposed architecture.\vspace{-0.5em}

\paragraph{Training}
AdamW~\cite{loshchilov2017decoupled} optimizer is used with the weight decay of 0.05 for our CvT-13, and 0.1 for our CvT-21 and CvT-W24. We train our models with an initial learning rate of 0.02 and a total batch size of 2048 for 300 epochs, with a cosine learning rate decay scheduler. We adopt the same data augmentation and regularization methods as in ViT~\cite{touvron2020training}. Unless otherwise stated, all ImageNet models are trained with an $224 \times 224$ input size. \vspace{-0.5em} 

\paragraph{Fine-tuning}
We adopt fine-tuning strategy from ViT~\cite{touvron2020training}. SGD optimizor with a momentum of 0.9 is used for fine-tuning. As in ViT~\cite{touvron2020training}, we pre-train our models at resolution $224\times224$, and fine-tune at resolution of $384\times384$. We fine-tune each model with a total batch size of 512, for 20,000 steps on ImageNet-1k, 10,000 steps on CIFAR-10 and CIFAR-100, and 500 steps on Oxford-IIIT Pets and Oxford-IIIT Flowers-102.

\subsection{Comparison to state of the art}
\begin{table*}[t]
    \centering
    {\small 
    
\begin{tabular}{@{\ }l|l|r|rr@{\ }|c|c@{\ }|c@{\ }}
    \toprule[1.2pt]
            & &   \#Param.  &         image &  FLOPs         &  \multicolumn{1}{c}{ImageNet} & \multicolumn{1}{|c|}{Real}& \multicolumn{1}{c}{V2}\\
    Method Type & Network &  \multicolumn{1}{c|}{(M)} & size & (G) &  top-1 (\%) &  top-1 (\%) &  top-1 (\%)\\
    \toprule[1.2pt]
    \multirow{3}{1.6in}{\emph{Convolutional Networks}} 
    & ResNet-50~\cite{he2016deep} & 25 & $224^{2}$ & 4.1 &   76.2 & 82.5 & 63.3\\
    & ResNet-101~\cite{he2016deep} &45 & $224^{2}$ & 7.9 & 77.4 & 83.7 & 65.7\\
    & ResNet-152~\cite{he2016deep} &60 & $224^{2}$ &11   & 78.3& 84.1 & 67.0 \\
    \toprule
    
    \multirow{12}{1.6in}{\emph{Transformers}}
    & ViT-B/16~\cite{dosovitskiy2020image}&86  & $384^{2}$  &55.5   & 77.9 & 83.6 & -- \\
    & ViT-L/16~\cite{dosovitskiy2020image} &307 & $384^{2}$ & 191.1   & 76.5  & 82.2 & -- \\
    \cmidrule{2-8}
    & DeiT-S~\cite{touvron2020training}[arxiv 2020] &22 & $224^{2}$  & 4.6  & 79.8 & 85.7  & 68.5 \\
    & DeiT-B~\cite{touvron2020training}[arxiv 2020] &86 & $224^{2}$ & 17.6  & 81.8  & 86.7 & 71.5\\
    \cmidrule{2-8}
    & PVT-Small~\cite{wang2021pyramid}[arxiv 2021] &25& $224^{2}$  & 3.8 &79.8 & -- & --  \\
    & PVT-Medium~\cite{wang2021pyramid}[arxiv 2021]&44& $224^{2}$  & 6.7 &81.2 & -- & --  \\
    & PVT-Large~\cite{wang2021pyramid}[arxiv 2021] &61& $224^{2}$ & 9.8 &81.7 & -- & --  \\
    \cmidrule{2-8}
    & T2T-ViT$_{t}$-14~\cite{yuan2021tokens}[arxiv 2021]&22& $224^{2}$  & 6.1 & 80.7 & -- & -- \\
    & T2T-ViT$_{t}$-19~\cite{yuan2021tokens}[arxiv 2021]&39& $224^{2}$  & 9.8  &81.4 & -- & --  \\
    & T2T-ViT$_{t}$-24~\cite{yuan2021tokens}[arxiv 2021]&64& $224^{2}$  & 15.0 & 82.2 & -- & --  \\
    \cmidrule{2-8}
    & TNT-S~\cite{han2021transformer}[arxiv 2021]&24& $224^{2}$  & 5.2 &81.3   & -- & --\\
    & TNT-B~\cite{han2021transformer}[arxiv 2021]&66& $224^{2}$ & 14.1  &82.8   & -- & --\\
    \midrule
    \multirow{4}{1.5in}{\emph{Convolutional Transformers}}
    & \textbf{Ours:} CvT-13& 20 & $224^{2}$  & 4.5  & 81.6 & 86.7 & 70.4 \\
    & \textbf{Ours:} CvT-21 & 32 & $224^{2}$  &7.1  & 82.5 & 87.2  & 71.3\\
    & \textbf{Ours:} CvT-13$_{\uparrow384}$& 20 & $384^{2}$  & 16.3  & 83.0 & 87.9 & 71.9 \\
    & \textbf{Ours:} CvT-21$_{\uparrow384}$ & 32 & $384^{2}$  &24.9 & \textbf{83.3} & \textbf{87.7} & \textbf{71.9}\\
    \cmidrule{2-8}
    & \textbf{Ours:} CvT-13-NAS& 18 & $224^{2}$  & 4.1  & 82.2 & 87.5 & 71.3 \\
    \toprule[1.2pt]
    
    \multirow{1}{1.6in}{$\emph{Convolution Networks}_{22k}$}
    & BiT-M$_{\uparrow480}$~\cite{kolesnikov2019big}& 928 & $480^{2}$ & 837 & 85.4 & --& --\\
    \midrule
    \multirow{3}{1.6in}{$\emph{Transformers}_{22k}$ }
    & ViT-B/16$_{\uparrow384}$~\cite{dosovitskiy2020image}& 86  & $384^{2}$   &55.5  & 84.0 & 88.4 & -- \\
    & ViT-L/16$_{\uparrow384}$~\cite{dosovitskiy2020image} &307 & $384^{2}$  & 191.1  & 85.2  & 88.4 & -- \\
    & ViT-H/16$_{\uparrow384}$~\cite{dosovitskiy2020image} &632 & $384^{2}$  &  -- & 85.1  & 88.7 & -- \\
    \midrule
    \multirow{3}{1.6in}{$\emph{Convolutional Transformers}_{22k}$}
    & \textbf{Ours:} CvT-13$_{\uparrow384}$ & 20 & $384^{2}$ &  16& 83.3  & 88.7 & 72.9 \\
    & \textbf{Ours:} CvT-21$_{\uparrow384}$ & 32 & $384^{2}$ & 25&   84.9 & 89.8 & 75.6 \\
    & \textbf{Ours:} CvT-W24$_{\uparrow384}$& 277 & $384^{2}$  & 193.2& \textbf{87.7} & \textbf{90.6} & \textbf{78.8} \\
    \bottomrule[1.2pt]
    \end{tabular}

    }\vspace{-0.3em}
    \caption{Accuracy of manual designed architecture on ImageNet~\cite{deng2009imagenet}, ImageNet Real~\cite{beyer2020we} and ImageNet V2 matched frequency~\cite{recht2019imagenet}. Subscript $_{22k}$ indicates the model pre-trained on ImageNet22k~\cite{deng2009imagenet}, and finetuned on ImageNet1k with the input size of $384\times384$, except BiT-M~\cite{kolesnikov2019big} finetuned with input size of $480 \times 480$.} \vspace{-0.3em}
    \label{tab:sota}
\end{table*}

We compare our method with state-of-the-art classification methods including Transformer-based models and representative CNN-based models on ImageNet~\cite{deng2009imagenet}, ImageNet Real~\cite{beyer2020we} and ImageNet V2~\cite{recht2019imagenet} datasets in Table~\ref{tab:sota}. 

Compared to Transformer based models, CvT achieves a much higher accuracy with fewer parameters and FLOPs. CvT-21 obtains a 82.5\% ImageNet Top-1 accuracy, which is 0.5\% higher than DeiT-B with the reduction of 63\% parameters and 60\% FLOPs. When comparing to concurrent works, CvT still shows superior advantages. With fewer paramerters, CvT-13 achieves a 81.6\% ImageNet Top-1 accuracy, outperforming PVT-Small~\cite{wang2021pyramid}, T2T-ViT$_t$-14~\cite{yuan2021tokens}, TNT-S~\cite{han2021transformer} by 1.7\%, 0.8\%, 0.2\% respectively. 

Our architecture designing can be further improved in terms of model parameters and FLOPs by neural architecture search (NAS)~\cite{xiyang2020danas}. In particular, we search the proper stride for each convolution projection of key and value ($stride = 1, 2$) and the expansion ratio for each MLP layer ($ratio_{MLP} = 2, 4$). Such architecture candidates with FLOPs ranging from 2.59G to 4.03G and the number of model parameters ranging from 13.66M to 19.88M construct the search space. The NAS is evaluated directly on ImageNet-1k. The searched CvT-13-NAS, a bottleneck-like architecture with $stride = 2, ratio_{MLP} = 2$ at the first and last stages, and $stride = 1, ratio_{MLP} = 4$ at most layers of the middle stage, reaches to a 82.2\% ImageNet Top-1 accuracy with fewer model parameters than CvT-13. 

Compared to CNN-based models, CvT further closes the performance gap of Transformer-based models. Our smallest model CvT-13 with 20M parameters and 4.5G FLOPs surpasses the large ResNet-152 model by 3.2\% on ImageNet Top-1 accuracy, while ResNet-151 has 3 times the parameters of CvT-13.  

Furthermore, when more data are involved, our wide model CvT-W24* pretrained on ImageNet-22k reaches to \textbf{87.7}\% Top-1 Accuracy on ImageNet \emph{without extra data} (\eg JFT-300M), surpassing the previous best Transformer based models ViT-L/16 by 2.5\% with similar number of model parameters and FLOPs.
    
\subsection{Downstream task transfer}
We further investigate the ability of our models to transfer by fine-tuning models on various tasks, with all models being pre-trained on ImageNet-22k. Table~\ref{tab:downstream_tasks} shows the results. Our CvT-W24 model is able to obtain the best performance across all the downstream tasks considered, even when compared to the large BiT-R152x4~\cite{kolesnikov2019big} model, which has more than $3\times$ the number of parameters as CvT-W24. 

\begin{table}[t]
\setlength{\tabcolsep}{6pt}
\renewcommand{\arraystretch}{1.1}
\small
    \centering
    {
    \footnotesize

\begin{tabular}{l|r|cccc}
\toprule
Model & 

\begin{tabular}[c]{@{}c@{}}Param\\ (M)\end{tabular} 
&
\begin{tabular}[c]{@{}c@{}}CIFAR\\ 10\end{tabular} 
&
\begin{tabular}[c]{@{}c@{}}CIFAR\\ 100\end{tabular} 
& Pets & 
\begin{tabular}[c]{@{}c@{}}Flowers\\ 102\end{tabular}  \\
\midrule
BiT-M~\cite{kolesnikov2019big} & 928 & 98.91 & 92.17 & 94.46 & 99.30 \\
\midrule
ViT-B/16~\cite{dosovitskiy2020image} & 86 & 98.95 & 91.67 & 94.43 & 99.38   \\
ViT-L/16~\cite{dosovitskiy2020image} & 307& 99.16 &93.44 & 94.73 & 99.61   \\
ViT-H/16~\cite{dosovitskiy2020image} & 632& 99.27 & 93.82 & \textbf{94.82} & 99.51   \\
\midrule
\textbf{Ours:} CvT-13 & 20 & 98.83&91.11 & 93.25 & 99.50    \\
\textbf{Ours:} CvT-21 &  32& 99.16 & 92.88 & 94.03 & 99.62    \\
\textbf{Ours:} CvT-W24 & 277 & \textbf{99.39} & \textbf{94.09}  & 94.73 & \textbf{99.72}  \\
\bottomrule
\end{tabular}
    }
    \caption{Top-1 accuracy on downstream tasks. All the models are pre-trained on ImageNet-22k data}
    \label{tab:downstream_tasks}
\end{table}

\subsection{Ablation Study}
We design various ablation experiments to investigate the effectiveness of the proposed components of our architecture. First, we show that with our introduction of convolutions,  position embeddings can be removed from the model. Then, we study the impact of each of the proposed Convolutional Token Embedding and Convolutional Projection components. 

\paragraph{Removing Position Embedding} \label{abl:pe}
Given that we have introduced convolutions into the model, allowing local context to be captured, we study whether position embedding is still needed for CvT. The results are shown in Table~\ref{tab:pos_embed}, and demonstrate that removing position embedding of our model does not degrade the performance. Therefore, position embeddings have been removed from CvT by default. As a comparison, removing the position embedding of DeiT-S would lead to 1.8\% drop of ImageNet Top-1 accuracy, as it does not model image spatial relationships other than by adding the position embedding. This further shows the effectiveness of our introduced convolutions. Position Embedding is often realized by fixed-length learn-able vectors, limiting the trained model adaptation of variable-length input. However, a wide range of vision applications take variable image resolutions. Recent work CPVT~\cite{chu2021really} tries to replace explicit position embedding of Vision Transformers with a conditional position encodings module to model position information on-the-fly. CvT is able to completely remove the positional embedding, providing the possibility of simplifying adaption to more vision tasks without requiring a re-designing of the embedding. 

\begin{table}[t]
    \centering
    \setlength{\tabcolsep}{8pt}
\renewcommand{\arraystretch}{1.1}
    \small
    \begin{tabular}{c|l|c|l|c}
\toprule
Method & Model & 
 \begin{tabular}[c]{@{}c@{}}Param\\ (M)\end{tabular} 
&Pos. Emb. & \begin{tabular}[c]{@{}l@{}}ImageNet\\ Top-1 (\%)\end{tabular} \\
\midrule
a & DeiT-S &22& Default &  79.8\\
b & DeiT-S &22& N/A &78.0  \\
\midrule
c & CvT-13 &20&Every stage&  81.5\\
d & CvT-13 &20&First stage &  81.4\\
e & CvT-13 &20&Last stage & 81.4 \\
f & CvT-13 &20&N/A & 81.6\\
 \bottomrule
\end{tabular} 
    \caption{Ablations on position embedding.}
    \label{tab:pos_embed}
\end{table}

\vspace{-1em}

\paragraph{Convolutional Token Embedding}
We study the effectiveness of the proposed Convolutional Token Embedding, and Table~\ref{tab:ablation_conv_embed} shows the results. Table~\ref{tab:ablation_conv_embed}d is the CvT-13 model. When we replace the Convolutional Token Embedding with non-overlapping Patch Embedding~\cite{dosovitskiy2020image}, the performance drops 0.8\% (Table~\ref{tab:ablation_conv_embed}a v.s. Table~\ref{tab:ablation_conv_embed}d). When position embedding is used, the introduction of Convolutional Token Embedding still obtains 0.3\% improvement (Table~\ref{tab:ablation_conv_embed}b v.s. Table~\ref{tab:ablation_conv_embed}c). Further, when using both Convolutional Token Embedding and position embedding as Table~\ref{tab:ablation_conv_embed}d, it slightly drops 0.1\% accuracy. These results validate the introduction of Convolutional Token Embedding not only improves the performance, but also helps CvT model spatial relationships without position embedding. 

\begin{table}[t]
    \centering
    \setlength{\tabcolsep}{9.4pt}
\renewcommand{\arraystretch}{1.1}
\small
        \begin{tabular}{c|c|c|c|c}
\toprule
Method 
& \begin{tabular}[c]{@{}l@{}}Conv.\\Embed.\end{tabular} 
& \begin{tabular}[c]{@{}l@{}}Pos.\\ Embed.\end{tabular}
& \begin{tabular}[c]{@{}c@{}}\#Param\\ (M)\end{tabular} 
& \begin{tabular}[c]{@{}l@{}}ImageNet\\ top-1 (\%)\end{tabular} \\
\midrule
a &  &  &  19.5 & 80.7 \\
b &  & \cmark &  19.9 & 81.1 \\
c & \cmark & \cmark &  20.3 & 81.4 \\
d & \cmark &  &  20.0 & 81.6\\
\bottomrule
\end{tabular}
    \caption{Ablations on Convolutional Token Embedding.}
    \label{tab:ablation_conv_embed}
\end{table}
\vspace{-1em}

\paragraph{Convolutional Projection} \label{abl:conv_proj}

\begin{table}[t]
\centering
\setlength{\tabcolsep}{6.4pt}
\renewcommand{\arraystretch}{1.1}
\small
\begin{tabular}{c|c|c|c|c}
\toprule
 Method
 & \begin{tabular}[c]{@{}c@{}}Conv. Proj. KV.\\ stride \end{tabular}
 & \begin{tabular}[c]{@{}c@{}}Params\\ (M)\end{tabular} 
 & \begin{tabular}[c]{@{}c@{}}FLOPs\\ (G)\end{tabular} 
 & \begin{tabular}[c]{@{}c@{}}ImageNet\\ top-1 (\%)\end{tabular} \\
 \midrule
 a & 1 & 20&  6.55& 82.3 \\
 b &2 & 20 & 4.53 & 81.6 \\
 \bottomrule
\end{tabular}
\caption{Ablations on Convolutional Projection with different strides for key and value projection. Conv. Proj. KV.: Convolutional Projection for key and value. We apply Convolutional Projection in all Transformer blocks.}
\label{tab:ablation_proj_stride}
\end{table}

\begin{table}[t]
\small
    \centering
     \setlength{\tabcolsep}{9.3pt}
\renewcommand{\arraystretch}{1.1}
    \begin{tabular}{c|c|c|c|c}
\toprule
\multirow{2}{*}{Method} 
& \multicolumn{3}{c|}{Conv. Projection} 
& \multirow{2}{*}{\begin{tabular}[c]{@{}l@{}}Imagenet\\ top-1 (\%)\end{tabular}} \\
 & Stage 1 & Stage 2 & Stage 3  &  \\
\midrule
\multicolumn{1}{c|}{a} & & &  &  80.6\\
\multicolumn{1}{c|}{b} & \cmark & & &  80.8\\
\multicolumn{1}{c|}{c} & \cmark  & \cmark  &   & 81.0\\
\multicolumn{1}{c|}{d} & \cmark  & \cmark  & \cmark   & 81.6\\
\midrule
\multicolumn{1}{c|}{\#Blocks} & 1  & 2  & 10 & \\
\bottomrule
\end{tabular}
    \caption{Ablations on Convolutional Projection v.s. Position-wise Linear Projection. \cmark indicates the use of Convolutional Projection, otherwise use Position-wise Linear Projection.}
    \label{tab:ablation_proj}
\end{table}

First, we compare the proposed Convolutional Projection with different strides in Table~\ref{tab:ablation_proj_stride}. By using a stride of 2 for key and value projection, we observe a 0.3\% drop in ImageNet Top-1 accuracy, but with 30\% fewer FLOPs. We choose to use Convolutional Projection with stride 2 for key and value as default for less computational cost and memory usage.

Then, we study how the proposed Convolutional Projection affects the performance by choosing whether to use Convolutional Projection or the regular Position-wise Linear Projection for each stage. The results are shown in Table~\ref{tab:ablation_proj}. We observe that replacing the original Position-wise Linear Projection with the proposed Convolutional Projection improves the Top-1 Accuracy on ImageNet from 80.6\% to 81.5\%.
In addition, performance continually improves as more stages use the design, validating this approach as an effective modeling strategy.

\section{Conclusion}
In this work, we have presented a detailed study of introducing convolutions into the Vision Transformer architecture to merge the benefits of Transformers with the benefits of CNNs for image recognition tasks. Extensive experiments demonstrate that the introduced convolutional token embedding and convolutional projection, along with the multi-stage design of the network enabled by convolutions, make our CvT architecture achieve superior performance while maintaining computational efficiency. Furthermore, due to the built-in local context structure introduced by convolutions, CvT no longer requires a position embedding, giving it a potential advantage for adaption to a wide range of vision tasks requiring variable input resolution.

{\small
\bibliographystyle{ieee_fullname}
\bibliography{egbib}
}

\end{document}